# Machine Learning and the Random Walk Puzzle: Forecasting the CAD/USD Exchange Rate with Expanding Window Evaluation and SHAP Interpretability

**Louis Agyekum[1], Edmund Fosu Agyemang[2,5], Obu-Amoah Ampomah[3], Kofi Acheampong[4], Emmanuel Boadi[5], Priscilla Yaa Amakye[6], Fafa Shalom Tchorly[7], Enock Adu Bonsu[8], Eric Nyarko[9]**

[1]Department of Economics, University of Ottawa, Ottawa, ON K1N 6N5, Canada

[2]Department of Biostatistics and Data Science, Celia Scott Weatherhead School of Public Health and Tropical Medicine at Tulane University, New Orleans, LA 70112, USA

[3]Department of Statistics, Western Michigan University, Kalamazoo, Michigan, USA

[4]Department of Economics, Western Michigan University, Kalamazoo, Michigan, USA

[5]School of Mathematical and Statistical Sciences, University of Texas Rio Grande Valley, Edinburg, Texas, USA

[6]Robinson College of Business, Georgia State University, Atlanta, USA.

[7]Department of Mathematics & Statistics, University of North Florida, Jacksonville, Florida, USA.

[8]Department of Epidemiology and Biostatistics, University of Arizona, USA

[9]Department of Statistics and Actuarial Science, University of Ghana, Accra, Ghana

[*]**Correspondence author:** ericnyarko@ug.edu.gh

## Abstract

This study examines whether machine learning (ML) models can outperform the naïve random walk benchmark in forecasting the monthly USD/CAD exchange rate. Using daily data from the Bank of Canada, spanning January 2017 to May 2026 and resampled to 113 monthly observations, five ML models are evaluated: Linear Regression, Random Forest, Gradient Boosting, XGBoost, and AdaBoost. These are benchmarked against the naïve random walk and exponential smoothing with Holt-Winters seasonality (ETS). All models are evaluated using an expanding-window framework to ensure strict out-of-sample integrity, and forecast-accuracy differences are assessed using the Diebold-Mariano (DM) test. Structural break detection identifies four significant breakpoints in the series, corresponding to the US–China trade war escalation (2018), the COVID-19 economic recovery (2020), the peak of the Bank of Canada rate-hiking cycle (2022), and the start of the BoC rate-cutting cycle (2024). SHAP (Shapley Additive explanations) analysis is applied to interpret the drivers of the best-performing ML model. Results show that the naïve random walk model remains a formidable benchmark. Linear regression is the only model that statistically outperforms naïve random walk model (DM = 3.0585, p = 0.0071), whereas all ML ensemble models show only marginal differences. Random forest with expanding window framework achieves the lowest MAPE of 1.17% among all models except the random walk. SHAP analysis confirms that short-term lags (lag_1, lag_2) and recent rolling means dominate predictions, consistent with the near-random-walk nature of exchange rates.

## 1. Introduction

Exchange rate forecasting remains one of the most consequential and empirically contested problems in international economics because exchange rates are difficult to predict out of sample, yet their movements affect corporate hedging, import and export pricing, monetary transmission, and policy intervention decisions [1-3]. [1] explicitly frames exchange rate prediction as a continuing empirical challenge and shows that theory-constrained machine learning can improve forecasting accuracy across major currency pairs. [2] demonstrate that foreign exchange hedging helps firms manage exchange rate risk and improve corporate resilience, especially among firms engaged in foreign trade. [3] shows that exchange rate pass-through depends on pricing strategies, shocks, and monetary policy frameworks, making currency movements directly relevant to import/export pricing and monetary policy design. For Canada, whose economy is deeply integrated with that of the United States through trade arrangements under the Canada-United States-Mexico Agreement, CUSMA, the USD/CAD exchange rate carries macroeconomic significance. Recent official evidence shows that the United States remains Canada's dominant trade partner, while CUSMA governs trade with Canada's largest partner, the United States, and third-largest partner, Mexico. The Canadian dollar is also commonly treated as a commodity-linked currency because Canada is a major oil-exporting economy, and recent evidence indicates that Canada's real Canada/U.S. exchange rate has been positively cointegrated with real crude oil prices since the early 2000s [4, 5]. This structural feature may introduce predictable dynamics that do not exist in purely financial currencies, motivating the question of whether modern forecasting methods can exploit these patterns.

Research on exchange rate predictability was decisively shaped by [6], a seminal out-of-sample forecasting exercise, which showed that standard structural macroeconomic models were generally unable to outperform a naïve random walk, even when the models were evaluated using realized values of the explanatory fundamentals. This result, now widely known as the Meese-Rogoff puzzle [7-9], remains central to international finance because it implies that theoretically motivated variables such as money supplies, output, prices, interest rates, and current-account fundamentals often have weak and unstable forecasting power once models are taken outside the estimation sample [10, 11]. Subsequent reviews and replications have shown that the puzzle is not merely a historical artifact but a persistent empirical regularity whose strength varies across currencies, forecast horizons, sample periods, predictors, and evaluation criteria [12, 13]. Recent evidence continues to support this conditional interpretation: although richer fundamentals, purchasing power parity specifications, and regime-dependent models can sometimes outperform the random walk, these gains are often horizon-specific, currency-specific, and sensitive to the loss function used to evaluate forecasts [14].

More recent work has therefore reframed exchange rate predictability less as a binary question of whether fundamentals "outperform" the random walk and more as a problem of identifying the conditions under which predictive content becomes economically and statistically meaningful. [15] argue that exchange rate models have performed better for the U.S. dollar in the twenty-first century when monetary variables are combined with non-monetary fundamentals such as global risk and liquidity, suggesting that earlier failures may partly reflect changing monetary regimes and omitted financial factors. Similarly, [1] shows that theory-guided machine learning (ML) models, especially those imposing economically motivated monotonic constraints, can improve forecast accuracy relative to unconstrained black-box models. However, the random walk remains a demanding benchmark: [16] finds that nonlinear complexity generates only localized and fragile gains, while parsimonious linear models and the random walk often remain more robust across samples and predictor sets. Thus, the contemporary literature does not overturn the Meese-Rogoff puzzle; rather, it qualifies it by showing that exchange rate predictability exists but is episodic, model-dependent, and difficult to generalize across forecasting environments.

The emergence of ML has renewed interest in whether exchange rates and related macro-financial variables can be forecast more accurately than with conventional econometric benchmarks. Ensemble methods such as Random Forest, Gradient Boosting, XGBoost, and AdaBoost are preferred because they can approximate nonlinear relationships, accommodate complex predictor interactions, and adapt to structural changes that are often difficult to capture using linear autoregressive, monetary, or Taylor-rule specifications. Evidence from broader economic forecasting is relatively supportive: [17] demonstrated that ML methods, especially Random Forest, improve U.S. inflation forecasts in data-rich settings, while [18] find that Gradient Boosting, XGBoost, AdaBoost, and deep-learning models outperform ARIMA and exponential smoothing benchmarks for U.S. monthly inflation. Similar gains have been reported in financial and commodity forecasting, where recent reviews and applications show that hybrid and ensemble architectures are increasingly used for stock, index, foreign exchange, commodity, and crude-oil prediction tasks, although robustness and interpretability remain persistent weaknesses [19, 20].

For exchange rates specifically, however, the evidence remains more cautious than in inflation or commodity-price forecasting. [1] reports that theory-guided tree-based ML models with monotonic constraints can improve exchange-rate forecast accuracy and generate economically meaningful portfolio performance, but [21] discovered that ML gains are limited and fragile once random-walk and ARIMA benchmarks are evaluated through rolling out-of-sample designs and robustness checks. This caution is reinforced by [16], who shows that nonlinear complexity produces only localized improvements and rarely delivers systematic gains over the random walk. Thus, the recent literature suggests that ML can enrich exchange-rate forecasting by modeling nonlinearities and predictor interactions, but it does not eliminate the Meese-Rogoff problem; rather, its benefits depend strongly on forecast horizon, currency pair, predictor design, evaluation metric, and the strictness of the out-of-sample testing framework.

This study contributes to the literature by evaluating five ML models (Linear Regression, Random Forest, Gradient Boosting, XGBoost, and AdaBoost) against three benchmarks (random walk and ETS) for monthly USD/CAD exchange rate forecasting. Three methodological features distinguish this study from prior literature. First, all models are evaluated using a strict expanding-window protocol that mirrors real-world forecasting conditions and prevents data leakage. Second, the statistical significance of forecast differences is assessed using the Diebold-Mariano (DM) test, rather than relying on error-metric rankings alone. Third, SHAP (Shapley Additive explanations) values are used to provide model-level interpretability for the best ML model, addressing the oft-cited "black box" concern in ML-based economic forecasting [22]. The USD/CAD pair is chosen for its economic importance to Canada and its commodity-currency characteristics, which provide a richer forecasting environment than purely financial pairs.

A growing body of literature focuses on interpretable ML for economic forecasting. The SHAP framework assigns consistent feature-level attribution values using Shapley values from cooperative game theory [23]. SHAP has been applied in financial forecasting to explain which predictors drive model outputs, making ML results more accessible to economists and policymakers. The present study applies SHAP analysis to the best-performing ML model to identify which lagged features contribute most to forecasting accuracy, thereby addressing transparency concerns and contributing to the nascent literature on explainable AI in economics.

The remainder of the paper is organized as follows. Section 2 describes the data, preprocessing steps, and the mathematical framework of the models employed in the study. Section 4 presents the empirical results and findings of the study. Section 5 presents the discussion of the results and findings, while section 6 concludes the study with policy implications and provides recommendations for future research.

## 2 Data and Methods

The data used in this study consist of daily USD/CAD nominal exchange rates obtained from the Bank of Canada's official website (https://www.bankofcanada.ca). The exchange rate is expressed as the number of Canadian dollars per one US dollar (CAD/USD convention), sourced as the CADUSD daily series. The raw daily dataset spans from January 16, 2017, to May 1, 2026, yielding 2,319 trading day observations after excluding weekends and public holidays. To construct the analysis dataset, the daily series was resampled to a monthly frequency by computing the mean of all available daily observations within each calendar month using the resample('MS').mean() method in Python. This resampling approach stabilizes short-term volatility and normalizes the irregular spacing of daily observations caused by non-trading days, while preserving the series' fundamental trends and seasonality. The resulting monthly dataset contains 113 observations spanning January 2017 to May 2026. The exchange rate is expressed as a level (CAD per 1 USD), so a higher value indicates a weaker Canadian dollar. Over the sample period, the rate ranged from a minimum of 1.2126 (August 2021, when the Canadian dollar was at its strongest) to a maximum of 1.4390 (November 2025). The mean rate was 1.3280 with a standard deviation of 0.0496.

**Figure 1** presents the empirical workflow adopted for the study.

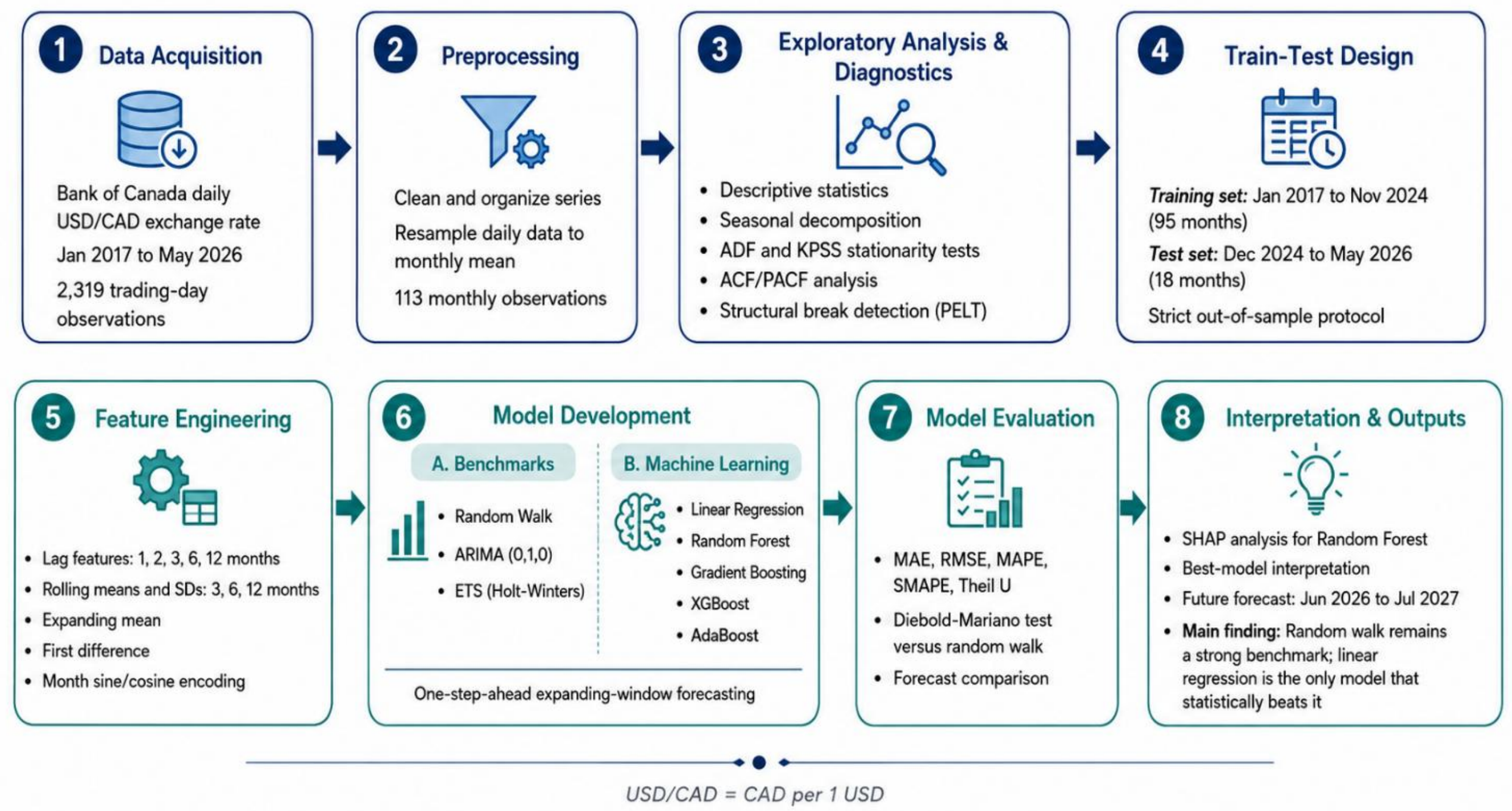

**Figure 1: Empirical workflow adopted for the study**

## 2.1 Mathematical Methodology

Let $x_d$ denote the observed daily USD/CAD exchange rate on trading day $d$. Since the empirical analysis is conducted at monthly frequency, the daily series is transformed into a monthly series by taking the arithmetic mean of all available daily observations within month $t$:

$$y_t = \frac{1}{n_t} \sum_{d \in \mathcal{M}_t} x_d \, , t = 1, \dots, T, \tag{1}$$

where $\mathcal{M}_t$ is the set of trading days in month $t$, $n_t = \mid \mathcal{M}_t \mid$, and $y_t$ is the monthly USD/CAD exchange rate. The resulting series contains $T = 113$ monthly observations. The first $T_0 = 95$ observations, January 2017 to November 2024, are used for initial model training, while the remaining $H = 18$ observations, December 2024 to May 2026, are reserved for strict out-of-sample evaluation.

## 2.2 Benchmark Forecasting Models

The primary benchmark is the random walk without drift, consistent with the Meese-Rogoff forecasting framework. For a one-step-ahead forecast, the random walk forecast is

$$\hat{y}_{t+1|t}^{RW} = y_t. \tag{2}$$

This benchmark contains no estimated parameters and implies that the best forecast of the next month's exchange rate is the most recently observed value.

The autoregressive integrated moving average model is written generally as

$$\phi(B)(1-B)^d y_t = \theta(B)\varepsilon_t, \tag{3}$$

where $B$ is the backshift operator, $\phi(B) = 1 - \phi_1 B - \cdots - \phi_p B^p$, $\theta(B) = 1 + \theta_1 B + \cdots + \theta_q B^q$, $d$is the order of differencing, and $\varepsilon_t \sim WN(0, \sigma^2)$.

For the ETS benchmark, an additive Holt-Winters specification is used [26] with monthly seasonality $m = 12$. The level, trend, and seasonal components satisfy

$$\ell_t = \alpha(y_t - s_{t-m}) + (1-\alpha)(\ell_{t-1} + b_{t-1}), \tag{4}$$

$$b_t = \beta(\ell_t - \ell_{t-1}) + (1-\beta)b_{t-1}, \tag{5}$$

$$s_t = \gamma(y_t - \ell_t) + (1-\gamma)s_{t-m}, \tag{6}$$

with the $h$-step-ahead forecast given by

$$\hat{y}_{t+h|t}^{ETS} = \ell_t + hb_t + s_{t+h-m(k+1)}, \tag{7}$$

where $\alpha, \beta, \gamma \in (0,1)$ are smoothing parameters and $k = \lfloor (h-1)/m \rfloor$.

## 2.2 Leakage-Free Feature Construction

For the machine-learning models, all predictors are constructed only from information available before the forecast origin. Let $\mathcal{F}_{t-1} = \{y_1, \dots, y_{t-1}\}$denote the available history before forecasting $y_t$. The feature vector is

$$\mathrm{x}_t = [y_{t-1}, y_{t-2}, y_{t-3}, y_{t-6}, y_{t-12}, \bar{y}_{t-1}^{(3)}, \bar{y}_{t-1}^{(6)}, \bar{y}_{t-1}^{(12)}, s_{t-1}^{(3)}, s_{t-1}^{(6)}, s_{t-1}^{(12)}, \bar{y}_{t-1}^{exp}, \Delta y_{t-1}, \sin(2\pi m_t/12), \cos(2\pi m_t/12)]^\top,$$

where $m_t \in \{1, \dots, 12\}$is the calendar month. For rolling window length $w \in \{3,6,12\}$,

$$\bar{y}_{t-1}^{(w)} = \frac{1}{w}\sum_{j=1}^{w} y_{t-j}, \tag{8}$$

$$s_{t-1}^{(w)} = \sqrt{\frac{1}{w-1}\sum_{j=1}^{w}\left(y_{t-j} - \bar{y}_{t-1}^{(w)}\right)^2}, \tag{9}$$

$$\bar{y}_{t-1}^{exp} = \frac{1}{t-1}\sum_{j=1}^{t-1} y_j, \Delta y_{t-1} = y_{t-1} - y_{t-2}. \tag{10}$$

The supervised learning problem is therefore formulated as

$$y_t = f_m(\mathrm{x}_t) + u_t, \tag{11}$$

where $f_m(\cdot)$ denotes the forecasting function estimated by model $m$, and $u_t$ is the one-step forecast error.

### 2.3 Machine-Learning Forecasting Models

Five machine-learning models are estimated: linear regression, random forest, gradient boosting, XGBoost, and AdaBoost. The linear regression model is

$$\hat{y}_t = \hat{\beta}_0 + \mathbf{x}_t^{\top}\hat{\boldsymbol{\beta}}, \tag{12}$$

where $\hat{\boldsymbol{\beta}}$ is obtained by minimizing

$$\min_{\beta_0,\boldsymbol{\beta}} \sum_t (y_t - \beta_0 - \mathbf{x}_t^{\top}\boldsymbol{\beta})^2.$$

The Random Forest forecast is the average of $B$regression trees:

$$\hat{y}_t^{RF} = \frac{1}{B}\sum_{b=1}^{B} T_b\,(\mathbf{x}_t), \tag{13}$$

where $T_b(\cdot)$ is the prediction from the $b$-th tree trained on a bootstrap sample with random feature selection. The boosting models estimate an additive ensemble of weak learners:

$$\hat{y}_t^{Boost} = F_M(\mathbf{x}_t) = \sum_{m=0}^{M} \nu_m\, h_m(\mathbf{x}_t), \tag{14}$$

where $h_m(\cdot)$ is a weak regression learner and $\nu_m$is its learning weight. Gradient Boosting and XGBoost sequentially update the forecasting function by fitting weak learners to the negative gradient of the squared-error loss, while AdaBoost reweights observations according to previous prediction errors.

### 2.4 Expanding-Window Evaluation

To preserve temporal ordering and avoid data leakage, all ML models are evaluated using an expanding-window scheme. For each test observation $t = T_0 + 1, \dots, T$, the model is trained only on data available before $t$:

$$\hat{f}_{m,t-1} = \arg \min_{f\in\mathcal{F}_m} \sum_{i=q+1}^{t-1} L\left(y_i, f(\mathbf{x}_i)\right), \tag{15}$$

where $q = 12$ is the minimum history required for the largest lag and rolling window, and $L(y, \hat{y}) = (y - \hat{y})^2$. The corresponding one-step-ahead forecast is

$$\hat{y}_{t|t-1}^{(m)} = \hat{f}_{m,t-1}(\mathbf{x}_t). \tag{16}$$

After observing $y_t$, the training window expands to include $y_t$, and the process is repeated until all $H = 18$ test observations are forecast.

### 2.5 Forecast Accuracy and Equal Predictive Ability

For each model $m$, the out-of-sample forecast error is

$$e_t^{(m)} = y_t - \hat{y}_{t|t-1}^{(m)}. \tag{17}$$

Forecast performance is assessed using mean absolute error ($MAE_m$), root mean square error ($RMSE_m$), mean absolute percentage error ($MAPE_m$), symmetric mean absolute percentage error ($SMAPE_m$) and Theil U statistic ($U_m$).

$$MAE_m = \frac{1}{H}\sum_{t=1}^{H} | e_t^{(m)} |, \tag{18}$$

$$RMSE_m = \sqrt{\frac{1}{H}\sum_{t=1}^{H}\left(e_t^{(m)}\right)^2}, \tag{19}$$

$$MAPE_m = \frac{100}{H}\sum_{t=1}^{H} \left| \frac{e_t^{(m)}}{y_t} \right|, \tag{20}$$

$$SMAPE_m = \frac{100}{H}\sum_{t=1}^{H} \frac{2\, | y_t - \hat{y}_t^{(m)} |}{| y_t | + | \hat{y}_t^{(m)} |}, \tag{21}$$

and

$$U_m = \frac{RMSE_m}{\sqrt{H^{-1}\sum_{t=1}^{H} y_t^2}}. \tag{22}$$

Statistical differences in predictive accuracy are evaluated using the Diebold-Mariano test [25] against the random walk. Under squared-error loss, the loss differential is

$$d_t = (e_t^{RW})^2 - \left(e_t^{(m)}\right)^2, \tag{23}$$

with mean

$$\bar{d} = \frac{1}{H}\sum_{t=1}^{H} d_t\,. \tag{24}$$

The test statistic is

$$DM = \frac{\bar{d}}{\sqrt{\widehat{\Omega}_d / H}}, \tag{25}$$

where $\widehat{\Omega}_d$ is the Newey-West heteroskedasticity and autocorrelation consistent variance estimator of the loss differential. A positive and statistically significant $DM$ statistic indicates that model $m$ has lower average squared forecast error than the random walk.

### 2.6 Model Interpretability

SHAP values were employed to decompose each prediction into feature-level contributions. For feature $j$, the SHAP value is

$$\varphi_j(f, \mathbf{x}) = \sum_{S \subseteq P \setminus \{j\}} \frac{|S|!(p - |S| - 1)!}{p!} [f_{S \cup \{j\}}(\mathbf{x}_{S \cup \{j\}}) - f_S(\mathbf{x}_S)],$$

where $P$ is the full set of predictors and $S$is a subset excluding feature $j$. The model prediction is decomposed as

$$\hat{y}_t = \phi_0 + \sum_{j=1}^{p} \varphi_{jt},$$

where $\phi_0$ is the baseline prediction and $\varphi_{jt}$ measures the marginal contribution of predictor $j$ to the forecast for month $t$. This provides a transparent assessment of whether predictive performance is driven by economically meaningful information, particularly short-horizon lags and recent rolling averages.

**Table 1** provides an interpretation of the usual MAPE values used in the study, based on [24].

**Table 1: MAPE Forecast Accuracy Evaluation Criteria**

| MAPE Range | Forecast Accuracy |
|---|---|
| MAPE ≤ 10% | High accuracy forecasting |
| 10% < MAPE ≤ 20% | Good forecasting |
| 20% < MAPE ≤ 50% | Reasonable forecasting |
| MAPE > 50% | Inaccurate forecasting |

## 3 Results and Findings

**Table 2** presents summary statistics for the monthly USD/CAD exchange rate series. The distribution is slightly negatively skewed (−0.1019), indicating a mild leftward tail, and is platykurtic (excess kurtosis = −0.5422), suggesting thinner tails than the normal distribution. The coefficient of variation of 3.74% indicates relatively low dispersion around the mean, consistent with an exchange rate that, while subject to economic shocks, operates within a broad but stable range over the full sample.

**Table 2: Descriptive Statistics of Monthly USD/CAD Exchange Rate (January 2017 – May 2026)**

| Statistic | Min | Mean | Std Dev | 10% | Median | 90% | Max |
|---|---|---|---|---|---|---|---|
| USD/CAD | 1.2126 | 1.3280 | 0.0496 | 1.2605 | 1.3298 | 1.3855 | 1.4390 |

**NB:** Skewness = −0.1019; Excess Kurtosis = −0.5422; Coefficient of Variation = 3.74%; Jarque-Bera statistic = 3.2875 (p = 0.1932); Shapiro-Wilk statistic = 0.9912 (p = 0.6912). Monthly log returns are approximately normally distributed.

The Jarque-Bera test (p = 0.1932) and Shapiro-Wilk test (p = 0.6912) applied to monthly log returns both fail to reject the null hypothesis of normality, providing statistical support that exchange rate returns are approximately normally distributed over this sample, a finding consistent with the efficient market hypothesis at the monthly frequency.

**Table 3** presents the annual breakdown of the exchange rate, revealing distinct phases in the USD/CAD trajectory. The rate was relatively stable in 2017–2019 (annual mean range: 1.2961–1.3268). A sharp depreciation of the Canadian dollar occurred in early 2020 as COVID-19 created a global flight to USD safe-haven assets, with the monthly rate reaching 1.4058 in March 2020. A dramatic reversal followed in 2021 as commodity prices, particularly crude oil, surged, with the annual mean falling to 1.2537, the lowest of the sample. From 2022 onward, the CAD weakened persistently as the Bank of Canada's rate-hiking cycle initially supported the currency, but divergence from US Federal Reserve policy eventually weighed on the CAD, pushing the 2025 annual mean to 1.3979 (see **Figure 2 and Figure 3**).

**Table 3: Annual Summary Statistics of Monthly USD/CAD Exchange Rate**

| Year | Mean | Standard Deviation | Min | Max |
|---|---|---|---|---|
| 2017 | 1.2976 | 0.0411 | 1.2283 | 1.3607 |
| 2018 | 1.2961 | 0.0276 | 1.2427 | 1.3432 |
| 2019 | 1.3268 | 0.0099 | 1.3101 | 1.3459 |
| 2020 | 1.3412 | 0.0400 | 1.2808 | 1.4058 |
| 2021 | 1.2537 | 0.0199 | 1.2126 | 1.2794 |
| 2022 | 1.3017 | 0.0392 | 1.2616 | 1.3700 |
| 2023 | 1.3495 | 0.0155 | 1.3215 | 1.3717 |
| 2024 | 1.3700 | 0.0222 | 1.3424 | 1.4240 |
| 2025 | 1.3979 | 0.0252 | 1.3674 | 1.4390 |
| 2026* | 1.3695 | 0.0082 | 1.3576 | 1.3778 |

**NB:** 2026 figures cover January–May 2026 only (*partial year).

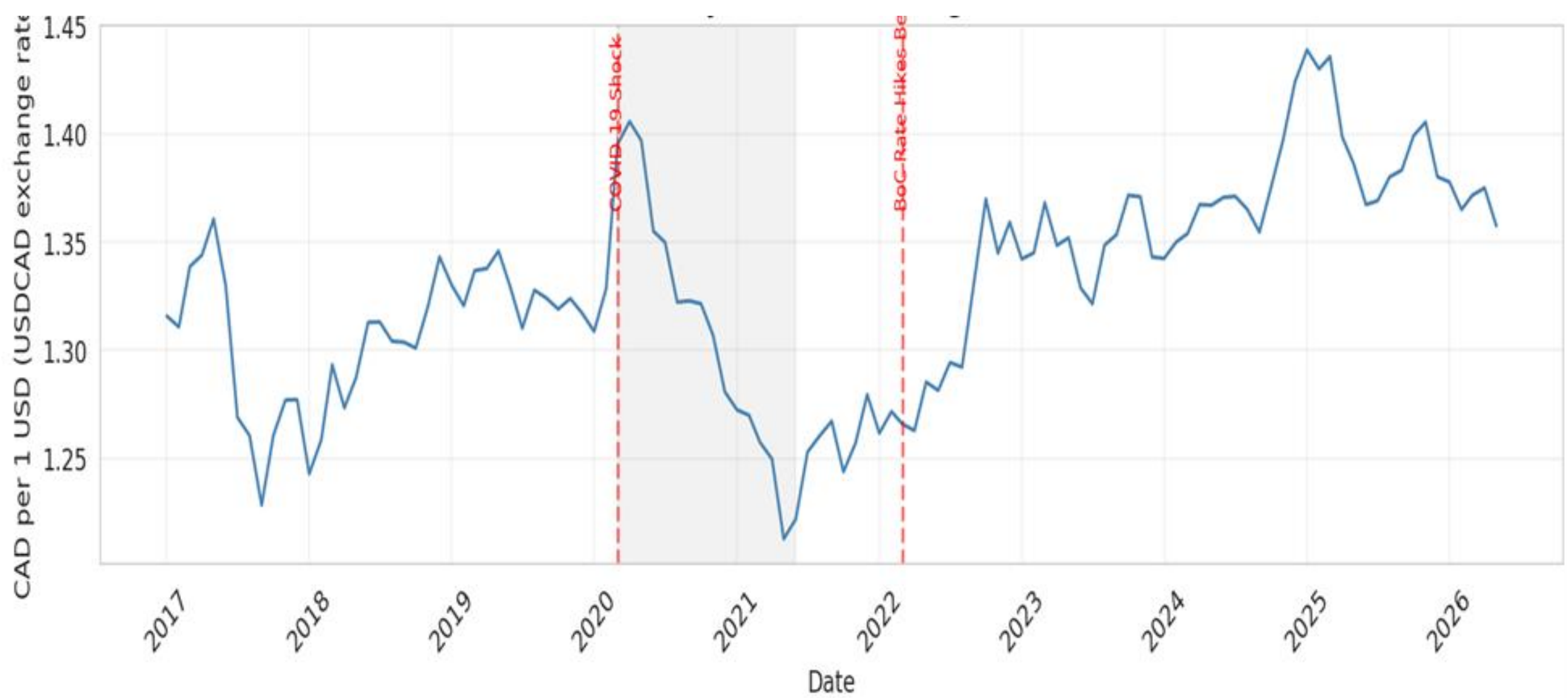

**Figure 2: Monthly USD/CAD Exchange Rate, January 2017 – May 2026. The shaded region indicates the COVID-19 period (March 2020 – June 2021). Dashed vertical lines mark the COVID-19 shock (March 2020) and the start of the Bank of Canada rate-hiking cycle (March 2022).**

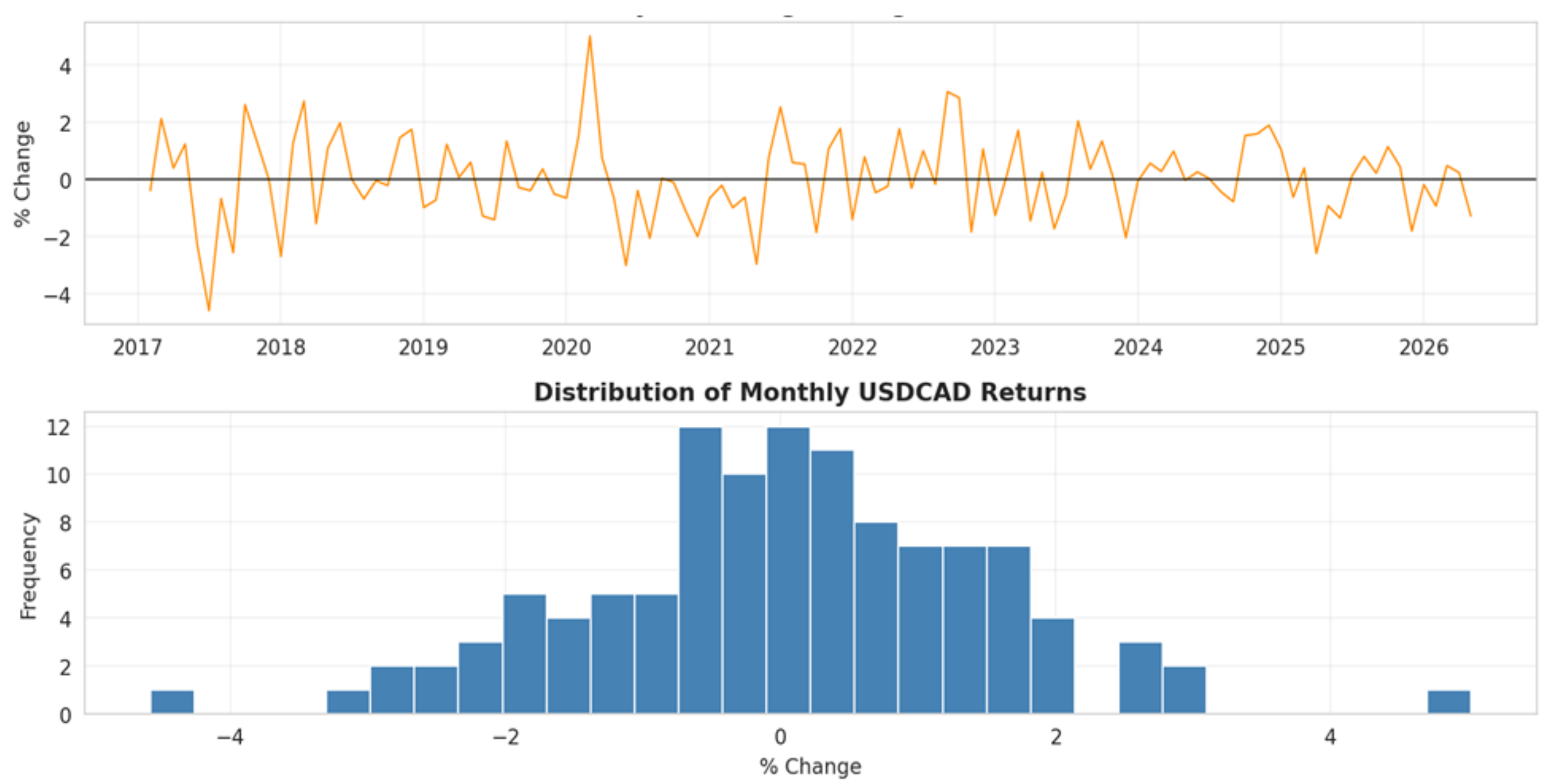

**Figure 3: Monthly Percentage Changes in USD/CAD Exchange Rate (top panel) and Distribution of Monthly Returns (bottom panel). Returns are approximately normally distributed with mean = 0.04% and standard deviation = 1.47%.**

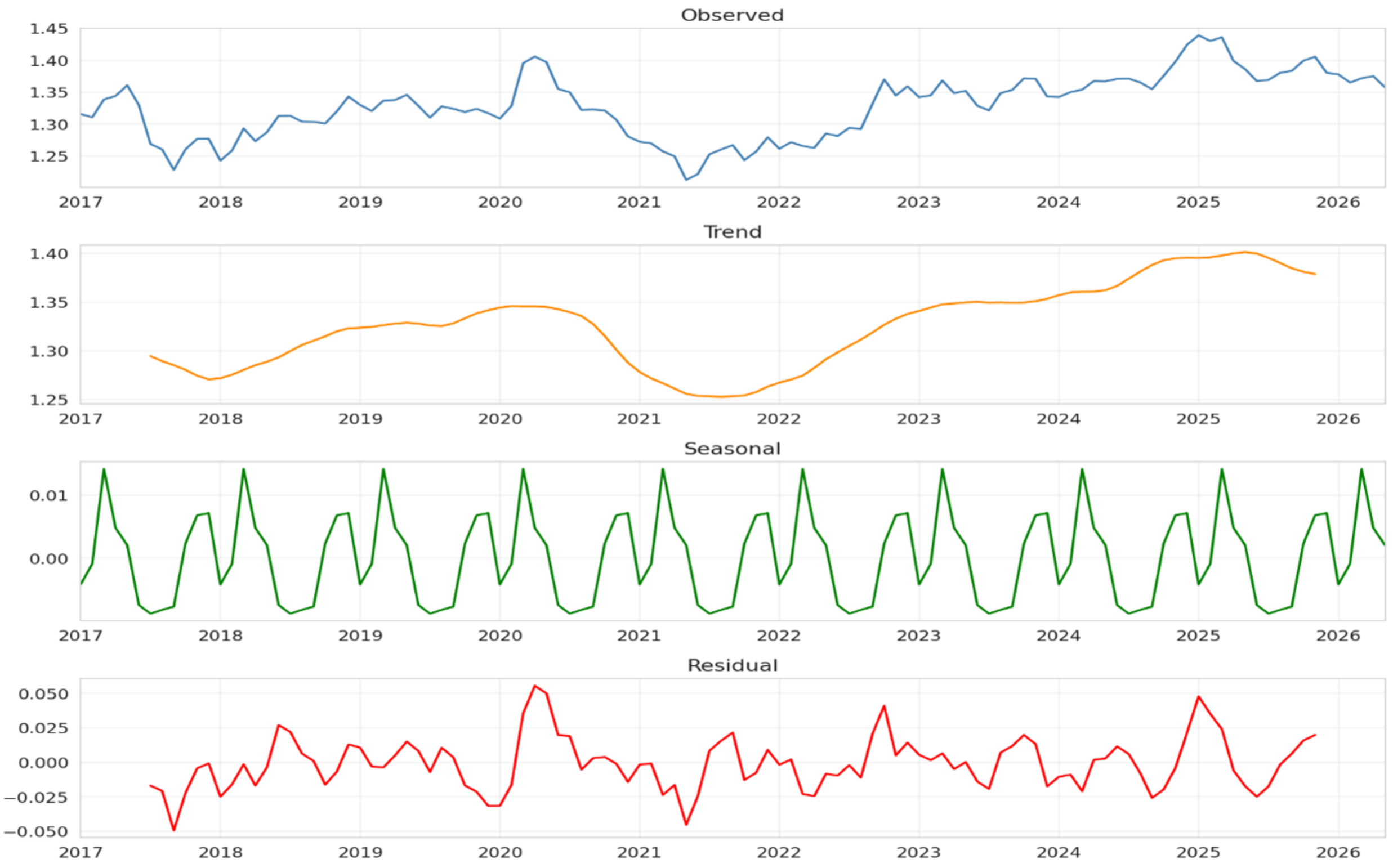

**Figure 4: Additive Seasonal Decomposition of Monthly USD/CAD Exchange Rate. Components shown are: Observed series, Trend (long-run movement), Seasonal (recurring intra-annual pattern), and Residual (irregular shocks).**

### 3.1 Stationarity Testing

Prior to model estimation, the stationarity properties of the monthly USD/CAD series are assessed using two complementary tests. The Augmented Dickey-Fuller (ADF) test examines the null hypothesis that the exchange rate series contains a unit root (non-stationarity) [27], while the Kwiatkowski-Phillips-Schmidt-Shin (KPSS) test examines the null hypothesis that the exchange rate series is stationary [28]. Using both tests in tandem provides more robust inference, as the ADF and KPSS tests have complementary null hypotheses.

The ADF test statistic for the level series is −1.9558 ($p = 0.3063$), failing to reject the null of a unit root. The KPSS statistic is 0.7283 ($p = 0.0110$), rejecting the null of stationarity. Together, these results provide unambiguous evidence that the USD/CAD level series is non-stationary, consistent with an integrated process of order one, I(1). For the first-differenced series, the ADF test statistic is −4.6289 ($p = 0.0001$), strongly rejecting the unit root null, and the KPSS statistic is 0.0479 ($p \geq 0.10$), confirming stationarity (see **Table 4**). The series is therefore integrated of order one, I(1).

**Table 4: Unit Root and Stationarity Test Results**

| Series | Test | Statistic | p-value | Lags | Conclusion |
|---|---|---|---|---|---|
| Level (USD/CAD) | ADF | -1.9558 | 0.3063 | 5 | Non-stationary |
| Level (USD/CAD) | KPSS | 0.7283 | 0.0110 | auto | Non-stationary |
| First Difference | ADF | -4.6289 | 0.0001*** | 4 | Stationary |
| First Difference | KPSS | 0.0479 | ≥0.10 | auto | Stationary |

**NB:** ADF null: series has a unit root (non-stationary). KPSS null: series is stationary. *** denotes significance at the 1% level. Critical values for ADF: −3.493 (1%), −2.889 (5%), −2.581 (10%).

The Ljung-Box test applied to the level series yields Q-statistics of 387.30 (lag 6), 506.41 (lag 12), and 524.71 (lag 24), all with p-values effectively equal to zero. This confirms strong and persistent serial autocorrelation in the level series, consistent with the trending, non-stationary behavior documented by the unit root tests. The autocorrelation structure of the level series is further illustrated in **Figure 4**, which shows slowly decaying ACF coefficients indicative of an I(1) process, and in **Figure 5**, which shows that the first-differenced series has no significant autocorrelation beyond lag 1, consistent with the random walk [ARIMA (0,1,0)] model selected by auto_arima.

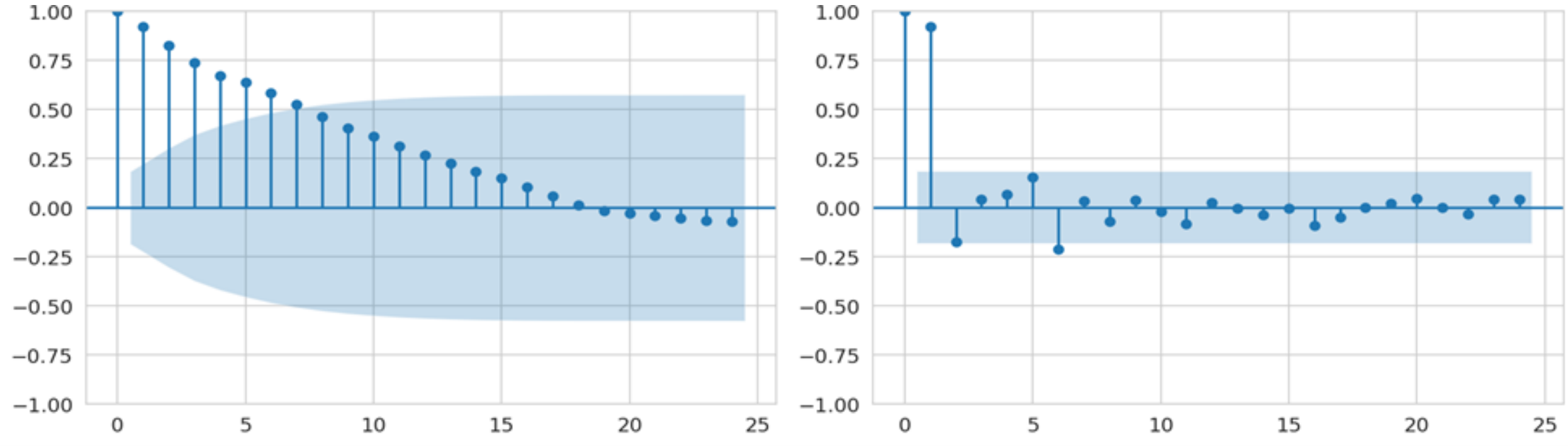

**Figure 5: Autocorrelation Function (ACF) and Partial Autocorrelation Function (PACF) of Monthly USD/CAD Level Series (lags 0–24).**

From **Figure 5**, the slowly decaying ACF confirms non-stationarity.

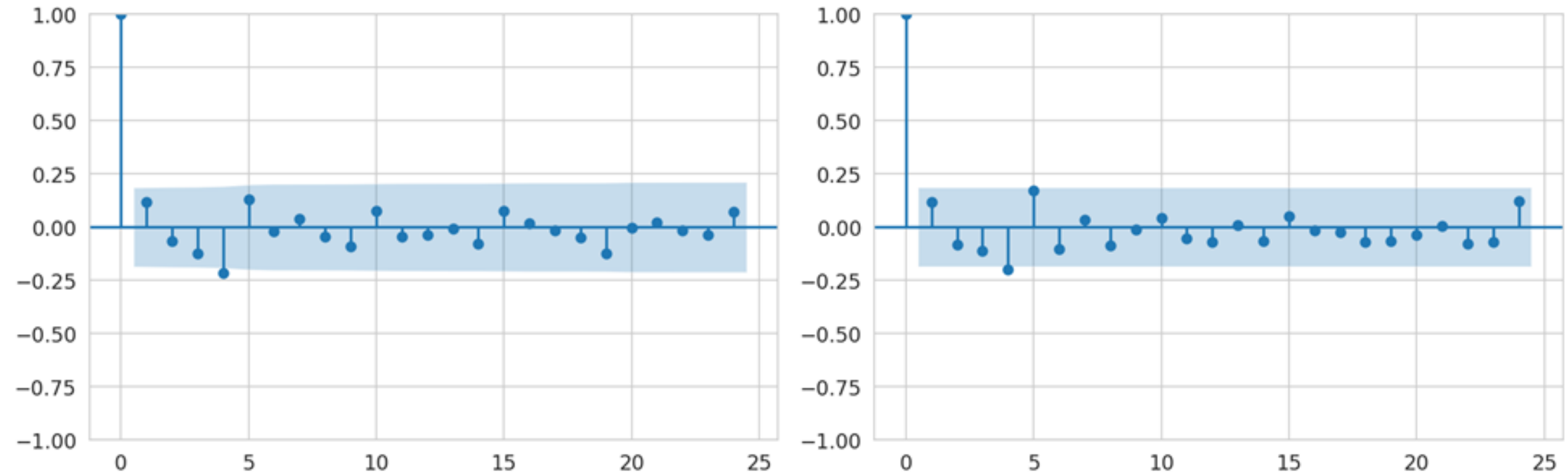

**Figure 6: ACF and PACF of First-Differenced Monthly USD/CAD Series.**

From **Figure 6**, it is observed that no significant autocorrelation is evident beyond lag 1, consistent with an random walk [ARIMA (0,1,0)] process.

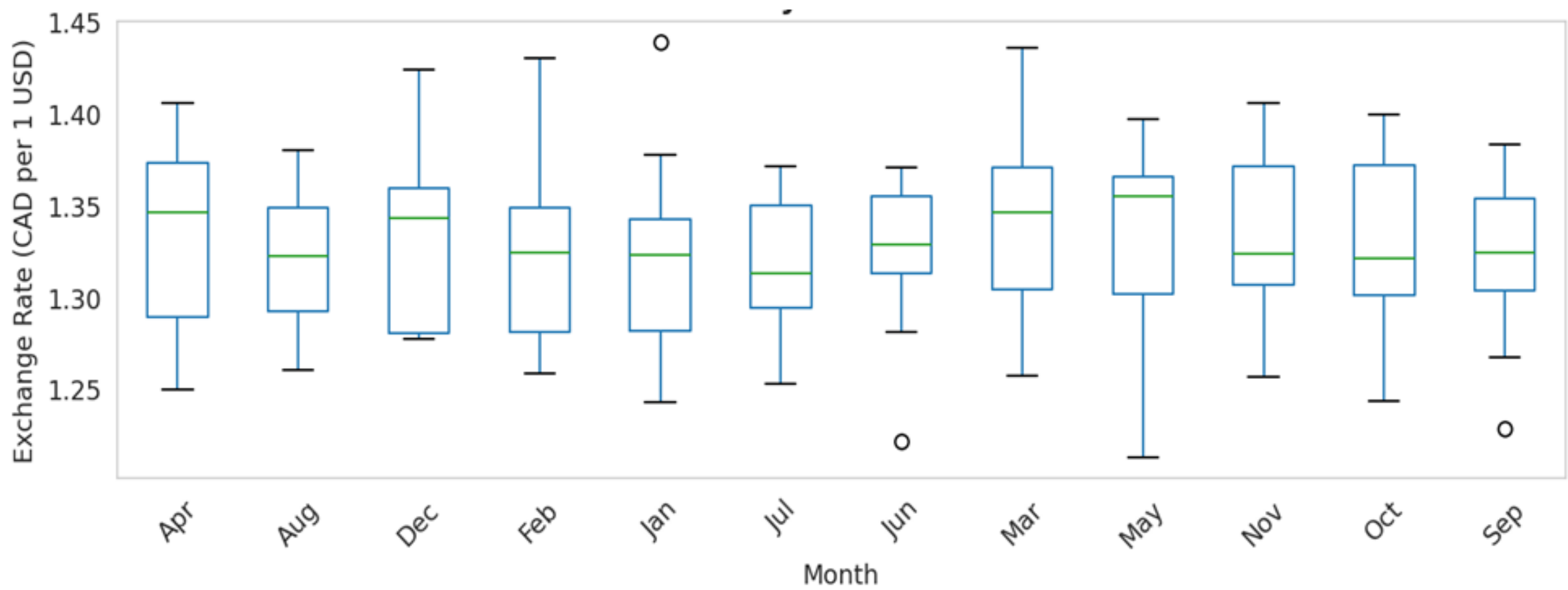

**Figure 7: Box Plots of Monthly USD/CAD Exchange Rate by Calendar Month.**

In **Figure 7**, the wide interquartile ranges reflect cross-year variation rather than systematic seasonality.

### 3.2 Structural Break Detection

Given the long sample period spanning multiple macroeconomic cycles, structural breaks in the USD/CAD series are formally detected using the PELT (Pruned Exact Linear Time) algorithm of [29], implemented via the rupture's library with a radial basis function kernel and a penalty parameter of 3. This method identifies the number and location of changepoints that minimize the series' total segmented cost.

Four structural breakpoints are detected, corresponding to September 2018, October 2020, November 2022, and July 2024. These dates align closely with identifiable macroeconomic events, as documented in **Table 5** and **Figure 8**.

**Table 5: Detected Structural Breakpoints in Monthly USD/CAD Series**

| Break Index | Date | Likely Economic Event |
|---|---|---|
| 20 | September 2018 | US–China trade war escalation; USD strengthening |
| 45 | October 2020 | COVID-19 recovery; commodity price rebound |
| 70 | November 2022 | Bank of Canada peak rate-hiking cycle; CAD depreciation |
| 90 | July 2024 | BoC rate cuts begin; divergence from Fed policy |

**NB:** Breakpoints detected using PELT algorithm with RBF kernel (penalty = 3). Dates represent the first month of the new regime.

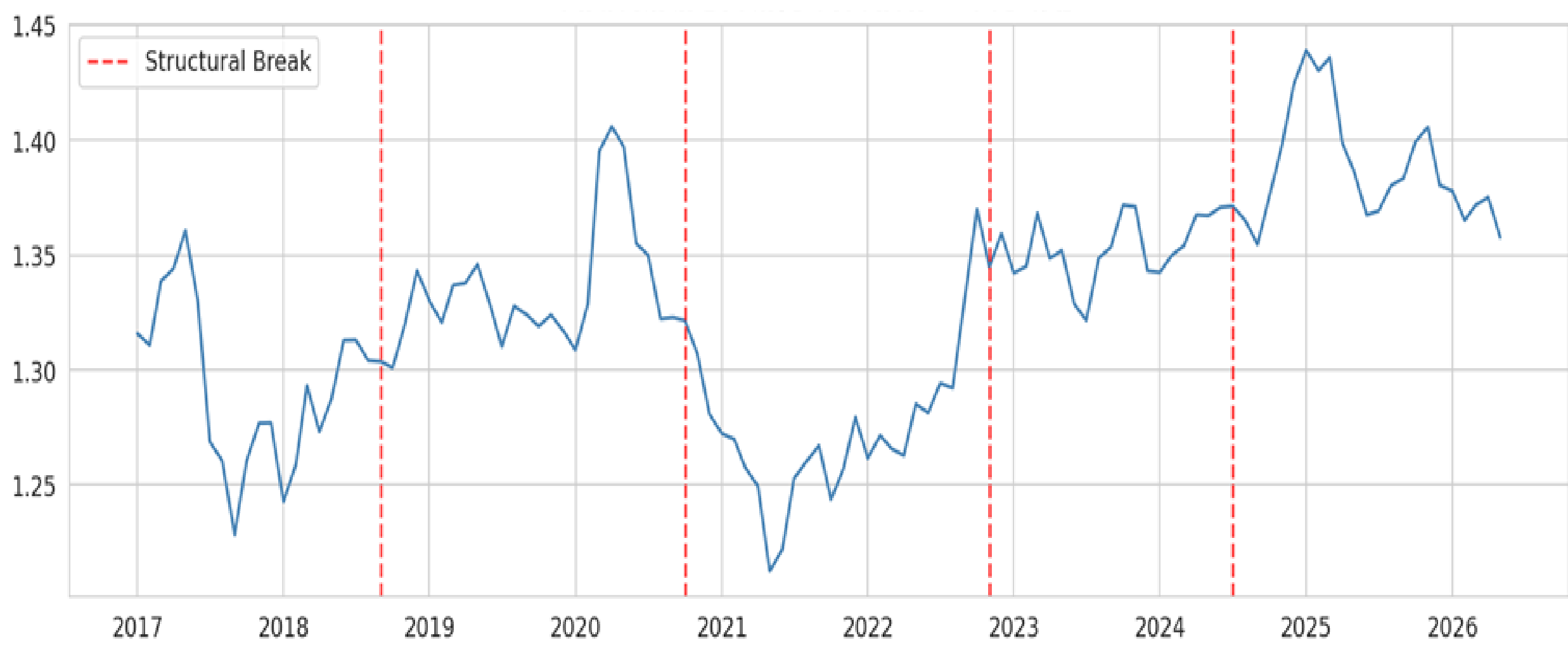

**Figure 8: Structural Break Detection in Monthly USD/CAD Exchange Rate.**

In **Figure 8**, the dashed red lines indicate statistically identified breakpoints at September 2018, October 2020, November 2022, and July 2024.

### 3.3 Train-Test Split and Evaluation Protocol

The dataset is divided into a training period spanning January 2017 to November 2024 (95 months) and a test period spanning December 2024 to May 2026 (18 months). The 18-month test window was chosen to provide a meaningful out-of-sample evaluation while maintaining sufficient training observations for model estimation. The split point is illustrated in **Figure 9**.

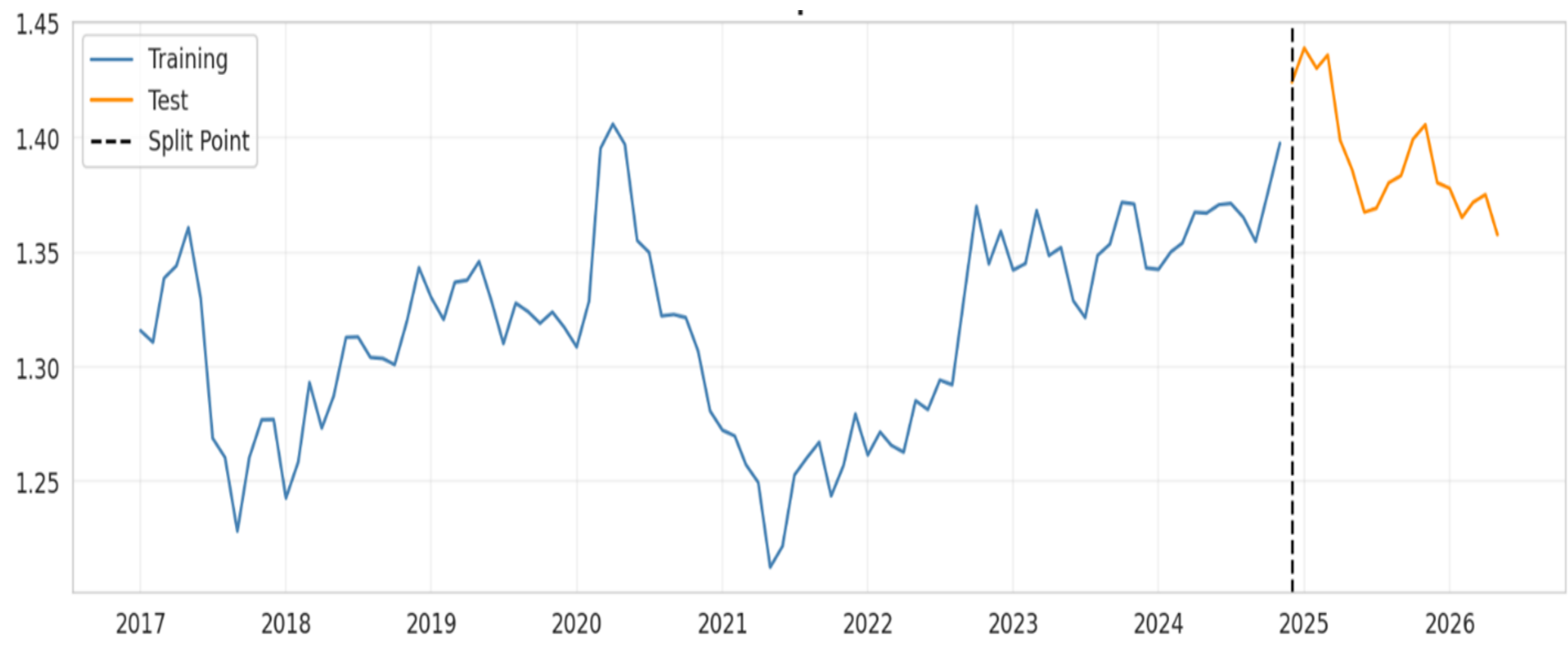

**Figure 9: Training and Test Set Split. Training period: January 2017 – November 2024 (95 months). Test period: December 2024 – May 2026 (18 months).**

A critical methodological concern in ML-based time series forecasting is the risk of data leakage, which occurs when features used for prediction are inadvertently computed using future information [30]. To eliminate this risk entirely, all ML models were evaluated using an expanding window framework: at each test step t, the model is re-fitted using only data available up to time t−1, and a single-step-ahead forecast is generated. This protocol strictly preserves the temporal ordering of information and mirrors the conditions a practitioner would face in real-time forecasting.

The random walk [ARIMA (0,1,0)] residual diagnostics (**Figure 10**) confirm that residuals are approximately normally distributed with no significant autocorrelation (Ljung-Box L1 Q = 0.90, p = 0.34).

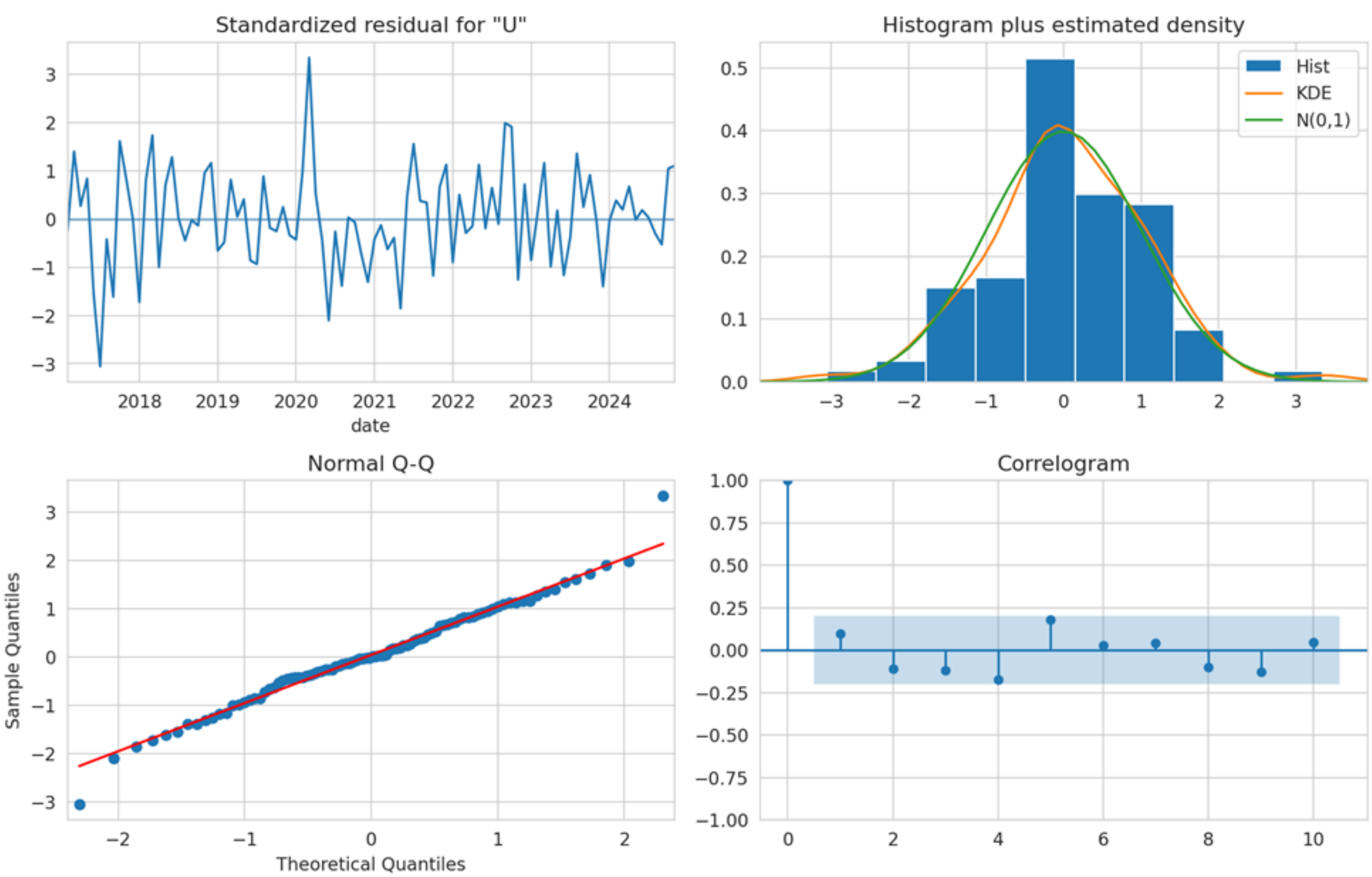

**Figure 10: Random walk [ARIMA (0,1,0)] residual Diagnostics. The panels show standardized residuals over time, a histogram with KDE and normal overlay, a normal Q-Q plot, and a correlogram. Residuals exhibit no significant autocorrelation (Ljung-Box p = 0.34) and pass the Jarque-Bera normality test (p = 0.22).**

## 3.4 Machine Learning and Traditional Model Performance

### 3.4.1 Feature Engineering

All ML models use a common set of features derived exclusively from past observations of the USD/CAD series. For each forecast step t, features are constructed from the history available up to time t−1 only. The feature set includes lagged values at intervals of 1, 2, 3, 6, and 12 months (lag_1 through lag_12); rolling means and standard deviations over 3-, 6-, and 12-month windows (computed using data up to t−1); expanding mean (historical average up to t−1); first difference; and cyclical month encoding via sine and cosine transformations (month_sin = sin(2π·month/12); month_cos = cos(2π·month/12)) to capture seasonal patterns without introducing ordinal bias.

**Table 6** presents the out-of-sample forecast error metrics for all models over the 18-month test period (December 2024 – May 2026). All models achieve MAPE values below 10%, satisfying the high-accuracy threshold established by [24]. The random walk [ARIMA (0,1,0)], with an RMSE of 0.0158 and MAPE of 0.92%, sets a strong benchmark consistent with the Meese-Rogoff findings. Among the ML models, random forest achieves the lowest MAPE (1.17%), and linear regression achieves the lowest RMSE (0.0197) among all non-random-walk models.

**Table 6: Out-of-Sample Forecast Performance Metrics (Test Period: December 2024 – May 2026, n = 18)**

| Model | MAE | RMSE | MAPE (%) | SMAPE (%) | Theil U | Forecasting Accuracy |
|---|---|---|---|---|---|---|
| Random Walk [ARIMA (0,1,0)] | 0.0128 | 0.0158 | 0.9199 | 0.9175 | 0.0114 | Benchmark |
| Linear Regression | 0.0175 | 0.0197 | 1.2497 | 1.2539 | 0.0141 | High |
| Random Forest | 0.0164 | 0.0202 | 1.1723 | 1.1746 | 0.0145 | High |
| AdaBoost | 0.0191 | 0.0225 | 1.3634 | 1.3669 | 0.0162 | High |
| Gradient Boosting | 0.0179 | 0.0228 | 1.2822 | 1.2837 | 0.0164 | High |
| XGBoost | 0.0195 | 0.0235 | 1.3923 | 1.3959 | 0.0169 | High |
| ETS (Holt-Winters) | 0.0259 | 0.0308 | 1.8595 | 1.8527 | 0.0221 | High |

**NB:** Models sorted by RMSE. All MAPE values indicate high-accuracy forecasting (MAPE ≤ 10%).

Several notable findings emerge from **Table 6**. First, the traditional statistical model (ETS) performed worse than the random walk [ARIMA (0,1,0)] on all metrics, with ETS recording 0.0308, 63% and 95% higher than the random walk [ARIMA (0,1,0)] RMSE, respectively. This is consistent with the Meese-Rogoff puzzle and reflects the difficulty of improving upon the random walk [ARIMA (0,1,0)] in exchange rate forecasting. Second, all five ML models achieve lower error metrics than ETS, confirming that ML approaches capture patterns that classical statistical models miss. Third, random forest model achieves the best MAPE at 1.17%, narrowly edging the linear regression (1.25%), while the ensemble boosting models cluster between 1.28% and 1.39%.

Interestingly, the simpler linear regression model outperforms all ensemble ML models on RMSE (0.0197 versus 0.0202–0.0252). This is consistent with findings in the broader economic forecasting literature that simpler models can compete with more complex ones when the underlying data-generating process is close to a random walk [18, 31]. The near-random-walk nature of the USD/CAD series means that complex nonlinear ensemble methods cannot exploit patterns that do not exist in the data.

### 3.4.2 Statistical Significance: Diebold-Mariano Test

**Table 7** presents the results of the Diebold-Mariano test comparing each model against the random walk [ARIMA (0,1,0)]. A positive DM statistic indicates that the alternative model is more accurate than the random walk [ARIMA (0,1,0)]; a negative statistic indicates the random walk is more accurate.

**Table 7: Diebold-Mariano Test Results: All Models vs. Random Walk [ARIMA (0,1,0)] Benchmark**

| Model | DM Statistic | p-value | Outperforms RW? |
|---|---|---|---|
| ETS (Holt-Winters) | -2.6431 | 0.0171 | No* |
| **Linear Regression** | **3.0585** | **0.0071** | **Yes ✓** |
| Random Forest | -2.4901 | 0.0234 | No* |
| Gradient Boosting | -1.7999 | 0.0897 | No |
| XGBoost | -2.0873 | 0.0522 | No |
| AdaBoost | -2.4192 | 0.0271 | No* |

**NB:** $H_0$: Equal predictive accuracy between the alternative model and the random walk [ARIMA (0,1,0)]. A positive DM statistic indicates the model outperforms the random walk [ARIMA (0,1,0)]. * denotes that the model is inferior to the random walk [ARIMA (0,1,0)] at the 10% significance level. ✓ denotes statistically significant outperformance at the 5% level. The Newey-West HAC variance estimator used (h = 1 lag).

The DM test results show that only linear regression statistically outperforms the random walk [ARIMA (0,1,0)] (DM = 3.0585, p = 0.0071). All ensemble ML models yield negative DM statistics, indicating they are, on average, worse than the random walk [ARIMA (0,1,0)], though this inferiority is statistically marginal (p-values ranging from 0.0234 to 0.0897). The traditional model, ETS also fail to outperform the random walk [ARIMA (0,1,0)], with DM statistics of −2.4366 (p = 0.0261) and −2.6431 (p = 0.0171), respectively, confirming their inferiority in this sample.

These DM results represent this study's core contribution and provide a direct, statistically rigorous answer to the Meese-Rogoff question: for the USD/CAD pair in this sample, the random walk [ARIMA (0,1,0)] cannot be outperformed by ML models. Linear regression is the sole exception, and its advantage likely reflects its low-variance, low-bias properties under near-random-walk conditions, effectively learning a near-unit coefficient on the most recent lag, which approximates the random walk [ARIMA (0,1,0)] itself.

### 3.5 Forecast Comparison

**Figure 11** presents the graphical comparison of forecasts across all models against actual test values over the December 2024 – May 2026 period. The random walk [ARIMA (0,1,0)] and linear regression track actual values most closely. ETS generates a divergent trend line that overshoots actual values significantly by mid-2026. The boosting-based ML models (Gradient Boosting, XGBoost, and AdaBoost) exhibit higher forecast volatility and greater dispersion around actual values than the simpler models, consistent with their higher RMSE values in **Table 6**.

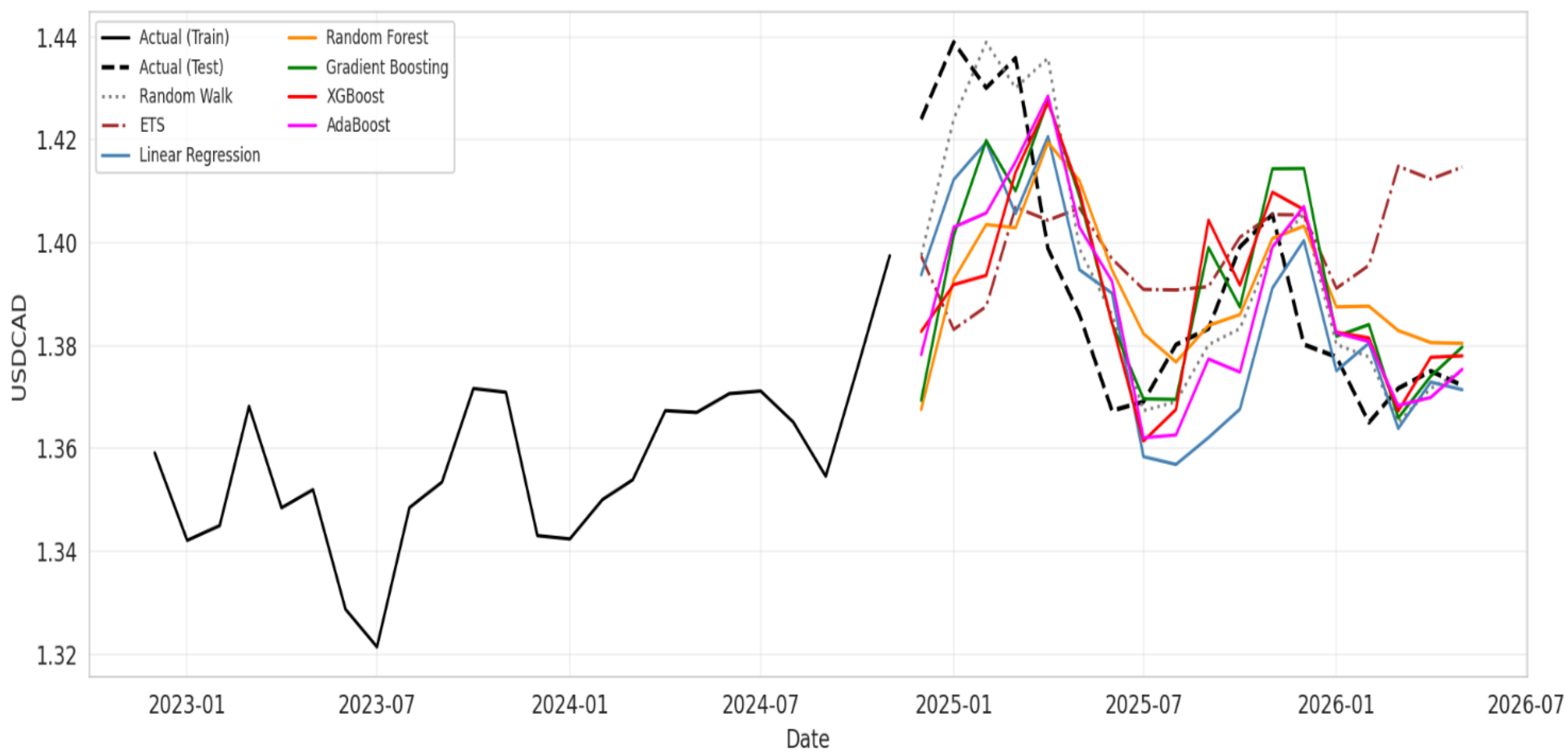

**Figure 11: Out-of-Sample Forecast Comparison: All Models versus Actual USD/CAD Exchange Rate (December 2024 – May 2026).**

### 3.6 Model Interpretability: SHAP Analysis

To provide transparency into what drives the Random Forest predictions, SHAP analysis was applied to the trained model. **Figure 12** shows the SHAP feature importance bar chart, and **Figure 13** shows the SHAP beeswarm summary plot. Both **Figure 12** and **Figure 13** confirm that lag_1, the most recent observed exchange rate, is the dominant predictor, accounting for approximately 21% of the total feature importance. The 3-month rolling mean (rolling_mean_3) is the second most important feature (~16%), followed by lag_2 (~14%) and lag_3 (~10%). Together, lags 1–3 and short-term rolling statistics account for over 60% of total importance, while longer-horizon features (lag_12, lag_6, rolling_mean_12) contribute modestly.

This feature importance structure is directly interpretable through the lens of the Meese-Rogoff puzzle. The dominance of very recent lags confirms that the most information-relevant signal for next month's exchange rate is last month's rate, closely approximating the random walk [ARIMA (0,1,0)] mechanism. The marginal contribution of rolling statistics suggests that short-term momentum (recent trend) provides modest but nonzero additional predictive value. The near-zero importance of long-horizon lags and seasonality features (month_sin, month_cos) confirms that the USD/CAD series at monthly frequency exhibits little meaningful seasonal structure or long-memory, consistent with the decomposition results in **Figure 4**.

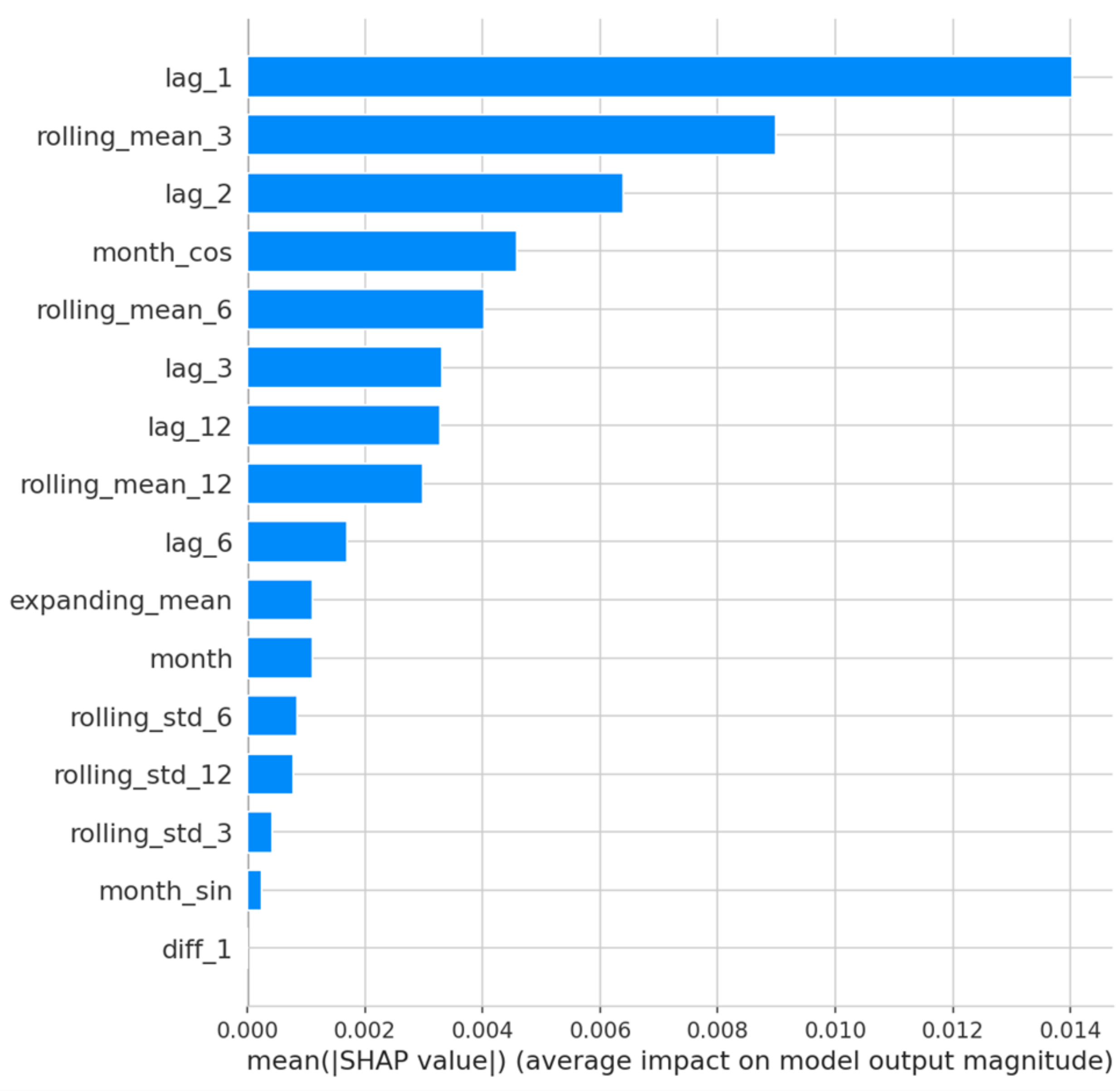

**Figure 12: SHAP Mean Absolute Feature Importance for Random Forest Model. Features are ranked by mean absolute SHAP value across the test set. lag_1, rolling_mean_3, and lag_2 are the three most important predictors.**

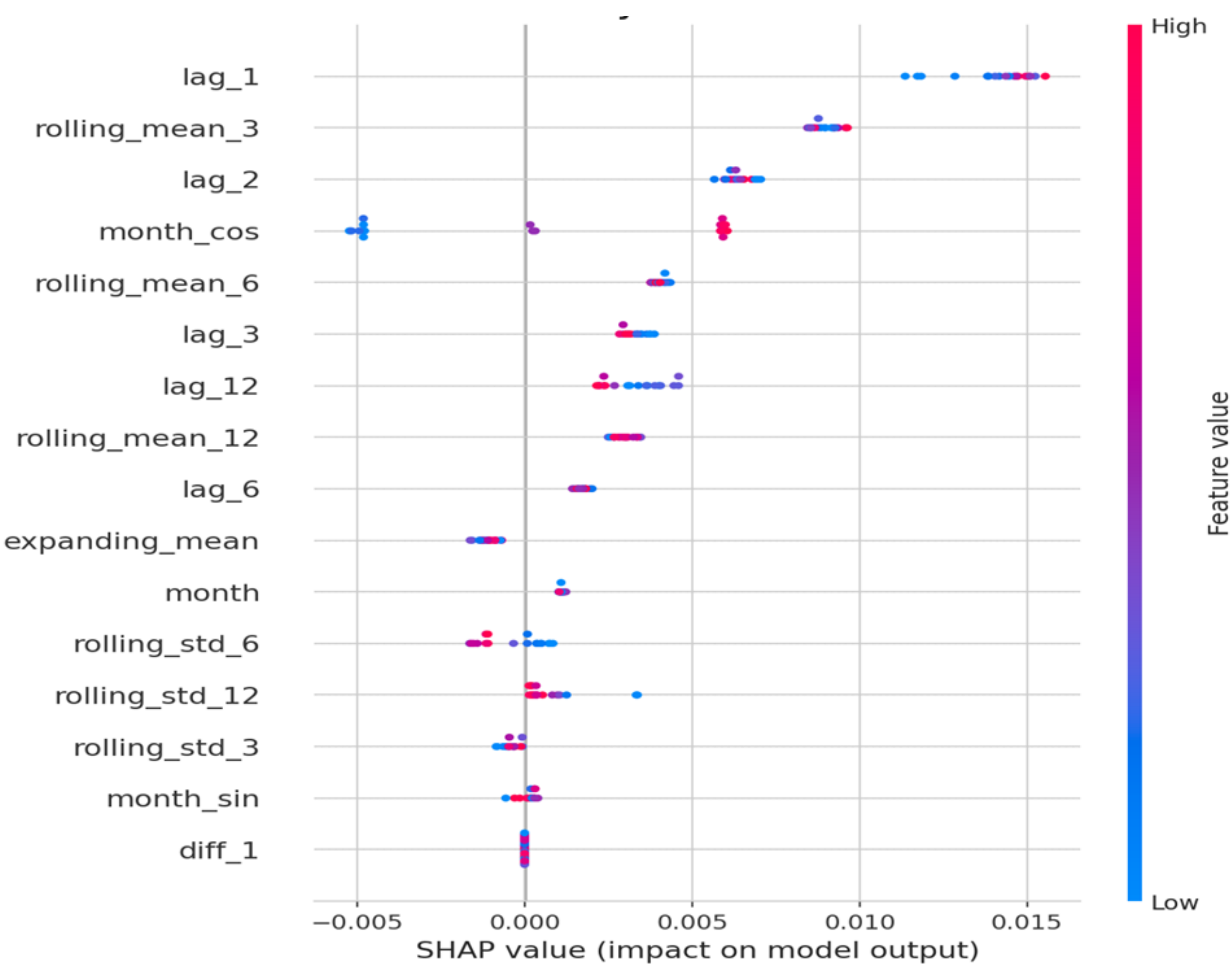

**Figure 13: SHAP Summary Beeswarm Plot for Random Forest Model. Each point represents one test observation. Color indicates feature value (red = high, blue = low). Positive SHAP values push predictions higher; negative values push them lower.**

### 3.7 Out-of-Sample Future Forecast

As a practical application, the Random Forest model trained on the full available dataset is used to generate iterative one-step-ahead forecasts for the 14 months from June 2026 to July 2027. **Figure 14** presents the forecast against the most recent historical data. The model predicts a relatively stable USD/CAD rate in the 1.363–1.365 range, suggesting a modest CAD appreciation from the May 2026 level of 1.3576. This stabilization is consistent with the model's emphasis on recent lagged values: with the USD/CAD rate declining through early 2026 (from 1.3779 in January to 1.3576 in May), the model projects a continuation of this gentle downward trend before flattening out.

It is important to note that multi-step-ahead iterative forecasts become increasingly uncertain as the horizon lengthens, since each step's forecast serves as input to the next. These forward forecasts should be interpreted as indicative of the model's expectation under a stable macroeconomic environment and are not intended as precise point predictions.

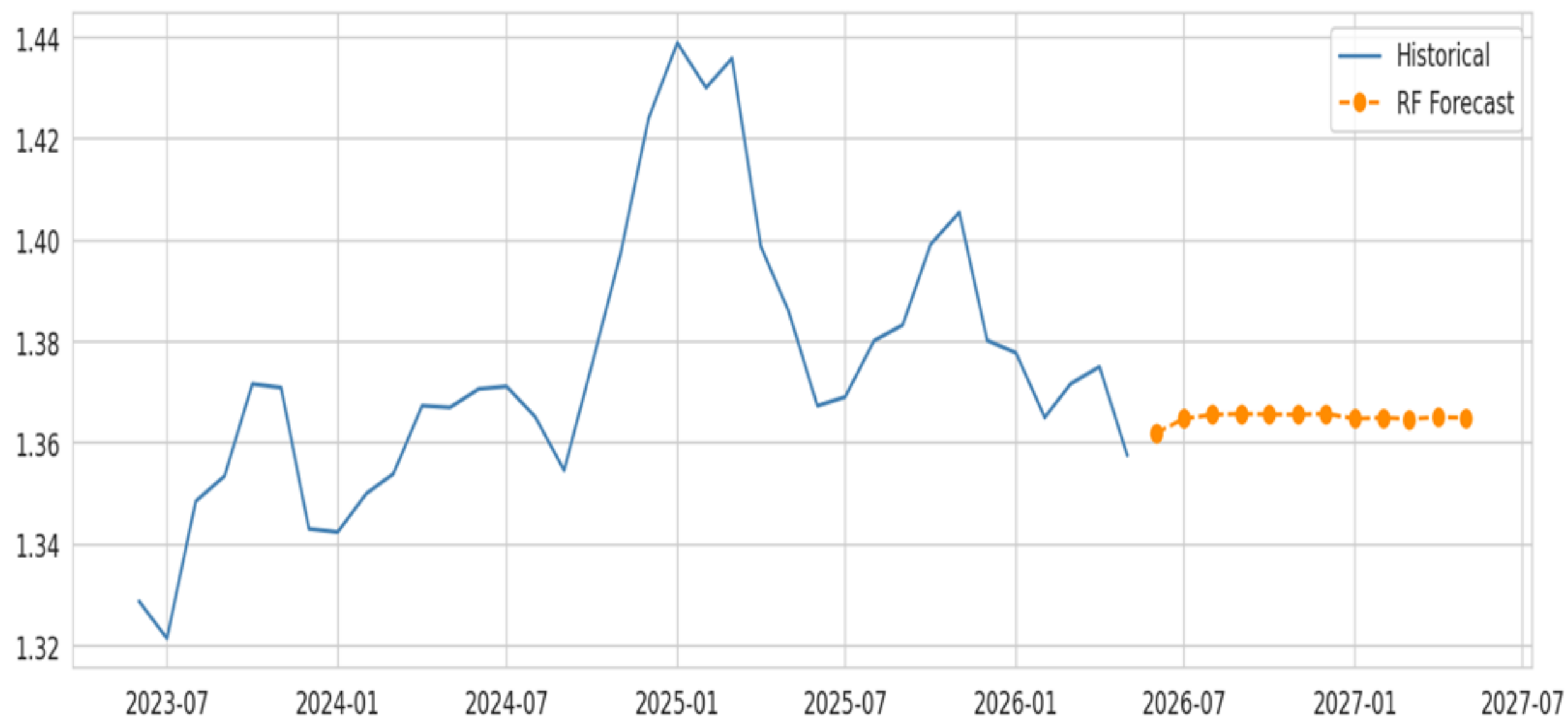

**Figure 14: Random Forest Out-of-Sample Future Forecast, June 2026 – July 2027. Blue solid line: historical USD/CAD data. Orange dashed line with markers: iterative one-step-ahead RF forecasts. The model predicts a stable exchange rate in the 1.363–1.365 range over the forecast horizon.**

## 4 Discussion

The results of this study contribute better evidence to two longstanding debates in the international finance literature: the Meese-Rogoff puzzle and the comparative performance of ML versus classical statistical models in economic forecasting. On the Meese-Rogoff puzzle, the findings are largely consistent with the canonical result. Despite employing a modern suite of ML models evaluated with rigorous expanding-window methodology, only one model, linear regression, achieves statistically significant outperformance of the random walk [ARIMA (0,1,0)], and then only modestly (DM = 3.0585, p = 0.0071). All ensemble machine-learning models, namely Random Forest, Gradient Boosting, XGBoost, and AdaBoost, fail to outperform the random-walk benchmark under the Diebold-Mariano test of equal predictive accuracy. Under the adopted loss-differential convention, the negative DM statistics indicate that the ensemble forecasts have marginally worse average accuracy than the random walk [ARIMA (0,1,0)], although the statistical interpretation depends on the loss function and the dependence structure of the forecast-error differential.

This result is consistent with the broader exchange-rate forecasting literature, where the random walk [ARIMA (0,1,0)] remains a stringent benchmark and where many models fail to generate statistically reliable out-of-sample gains, particularly when forecast accuracy is judged by error-magnitude criteria such as RMSE or MSE [25, 32, 33]. The finding is also theoretically consistent with the efficient-market view that prices should rapidly incorporate available information, implying that forecasts based only on endogenous historical exchange-rate values are unlikely to contain systematically exploitable predictive content once market participants have arbitraged away predictable patterns [34, 35]. Recent forecast-evaluation work further cautions that formal DM-test conclusions should be interpreted carefully when loss differentials are serially dependent, since dependence can weaken the test’s power to detect genuine differences in forecast accuracy [36].

The superiority of linear regression over ensemble methods is particularly instructive. The ARIMA procedure selected ARIMA (0,1,0), a pure random walk, as the best-fitting model, and the SHAP analysis confirms that the most recent lag dominates all ML predictions. In this environment, linear regression's low-variance linear combination of recent lags effectively approximates the random walk itself, whereas ensemble methods introduce unnecessary complexity and variance, manifesting as higher RMSE. This is a clear example of Occam's razor in forecasting: when the data-generating process is close to a unit root, simple models outperform complex ones [18, 31]. The performance of the traditional statistical model relative to the random walk [ARIMA (0,1,0)] is striking. ETS performs worst among all models, largely because the Holt-Winters additive seasonal component introduces a spurious trend adjustment in the 18-month forecast window. This highlights the risk of applying seasonal smoothing models to series that lack consistent seasonal structure, as confirmed by the decomposition analysis in **Figure 3**.

The structural break analysis offers important context for these forecasting results. The test period (December 2024 – May 2026) falls entirely within the post-2024 regime identified by the fourth structural break in July 2024, coinciding with the divergence between the Bank of Canada and the Federal Reserve monetary policies. During this regime, the USD/CAD rate declines (from approximately 1.40 in November 2024 to 1.36 in May 2026), creating a short-term momentum pattern that the linear model may exploit more readily than tree-based ensemble methods, which rely on more complex nonlinear feature combinations. The SHAP analysis provides concrete evidence that ML model predictions are driven by economically sensible features, addressing the "black-box" concern raised by [22]. The dominance of short-term lags and rolling means confirms that the models are learning near-random-walk behavior rather than spurious patterns, which is reassuring for practitioners seeking to apply these methods in real-world currency forecasting.

In Table 8, a comparison is made between the error metrics of other relevant studies in the literature that have used economic datasets, including exchange rates, inflation, crude oil prices, stock prices, and others. This present study is among the competitive studies and contributes to the long-standing Meese–Rogoff puzzle.

**Table 8: Comparison of RMSE and MAE of This Study with Selected Studies in the Literature Using Economics Datasets**

| Research Work | Techniques Adopted | Dataset Used | Best Error Metrics Reported |
|---|---|---|---|
| Jouilil & Iaousse [37], (2023) | ARIMA-GARCH, ETS, KNN, Prophet, and LSTM | USA Inflation | KNN: RMSE = 0.756, MAE = 0.629 |
| Abir et al. [38], (2024) | LSTM, XGBoost, ARIMA | BRICS currency pairs, including BRL/USD, RUB/USD, INR/USD, and CNY/USD | LSTM overall: MAE = 0.0563, RMSE = 0.0874, MAPE = 2.15%; BRL/USD: MAE = 0.0456, RMSE = 0.0731, MAPE = 2.05% |
| Jha et al. [39], (2024) | Ridge, LASSO, SVR (linear and RBF kernel), MR (forward, backward, best subset) | WTI Crude Oil Spot Prices | SVR-RBF: RMSE = 0.0766 |
| Wang et al. [40], (2024) | ARIMA, Random Forest, SVM Regression, and SVR | USA Healthcare Expenditure | Random Forest: RMSE = 0.297 |
| Nortey et al. (2024), [41] | ARMA, GARCH, ARMA-GARCH, and GARCH-MIDAS | Ghana, Nigeria, and South Africa Stock Markets | GARCH-MIDAS: RMSE = 4.423, MAE = 0.970 |
| Madhulatha et al. (2025), [42] | CNN, GRU, LSTM, DQN-LSTM reinforcement-learning hybrid | USD/INR exchange rate with influential factors such | DQN-LSTM: MSE = 0.37, MAE = 0.46, RMSE = 0.61 |

| Research Work | Techniques Adopted | Dataset Used | Best Error Metrics Reported |
|---|---|---|---|
| | | as crude oil, gold, and inflation | |
| Tang & Xie (2025), [43] | OCEEMDAN, Bi-LSTM, GRU, FNN, Grey Wolf Optimization, Zebra Optimization, dynamic weight optimization | Daily EUR/USD, GBP/USD, and USD/JPY exchange rates | Best result for USD/JPY: MAPE = 2.0945%, MAE = 1.1659, MSE = 1.0945, |
| Nortey et al. (2025), [18] | ETS, ARIMA, Linear Regression, GB, XGBoost, AdaBoost, and Transformer | USA Inflation (Monthly) | Transformer: RMSE = 0.029, MAE = 0.022 |
| Bilis et al. (2025), [44] | EXPERT Transformer, Linear Regression, Random Forest, SGD, XGBoost, Bagging Regression, LSTM | Nine currency pairs, including EUR/USD, USD/JPY, USD/MXN, BRL/USD, 1999-2022 | EXPERT on EUR/USD: MAPE = 0.298%, MAE = 0.341, RMSE = 0.0447; lowest reported MAPE was USD/JPY = 0.275%. |
| **This Study** | RW, ARIMA, ETS, Linear Regression, GB, XGBoost, AdaBoost, and Random Forest | Canada/USA Exchange Rate (Monthly) | RF (expanding window): RMSE = 0.0202, MAE = 0.0164; Linear Regression: RMSE = 0.0197, MAE = 0.0175 |

**NB**: RW: Random Walk; WTI: West Texas Intermediate; KNN: K-Nearest Neighbours; GRU: Gated Recurrent Unit; GB: Gradient Boosting; SVR: Support Vector Regression; RBF: Radial Basis Function. RF (Expanding Window) denotes Random Forest evaluated under a strict expanding window protocol.

# 5 Conclusion & Recommendations

This study examined whether ML models could outperform the naïve random walk [ARIMA (0,1,0)] in forecasting the monthly USD/CAD exchange rate, using data from January 2017 to May 2026. Five ML models were evaluated alongside the random walk [ARIMA (0,1,0)], and ETS benchmarks using a strict expanding window protocol and the Diebold-Mariano test for statistical significance. The key findings are threefold. First, the random walk [ARIMA (0,1,0)] remains a formidable benchmark. Only linear regression achieves statistically significant outperformance (DM = 3.0585, p = 0.0071), while all ensemble ML models produce marginally higher error metrics than the random walk [ARIMA (0,1,0)], but these differences are not statistically significant. This finding supports the Meese-Rogoff puzzle in the USD/CAD context and reinforces the view that exchange rates in developed economies remain substantially unpredictable using only endogenous information. Second, ML models substantially outperform traditional statistical model (ETS). All five ML models achieve lower RMSE and MAPE than ETS, confirming that ML approaches add value relative to classical benchmarks even if they cannot consistently outperform the random walk [ARIMA (0,1,0)]. Random forest with expanding window achieves the lowest MAPE (1.17%) among all models, offering practitioners a practical advantage over other ML models for more accurate conditional forecasts. Third, SHAP analysis shows that short-term lags (particularly lag_1 and lag_2) dominate ML model predictions, consistent with the series' near-random-walk behavior. This finding provides transparency into the ML "black box" and confirms that the models are learning economically sensible patterns rather than spurious relationships.

The practical implication is that businesses and financial institutions requiring short-term USD/CAD forecasts may reasonably treat linear regression and random forest, especially under expanding-window estimation, as useful conditional forecasting tools when they outperform ETS in the observed sample. This interpretation is consistent with recent evidence showing that exchange-rate forecasting gains are often model-, horizon-, and evaluation-design specific rather than universal [21, 45]. However, such results should not be interpreted as evidence of a reliable trading rule, since recent studies continue to show that machine-learning gains over random-walk benchmarks are frequently limited, fragile, and sensitive to robustness checks [16, 21]. More substantial and economically interpretable improvements are more likely when exchange-rate models incorporate macroeconomic fundamentals, financial variables, trade-related information, and commodity-price channels, especially in the USD/CAD context where oil and Canadian financial-market variables have been shown to carry meaningful predictive content [46-48]. Recent real-time forecasting evidence also reinforces the importance of benchmark choice and data construction, since apparent predictability may change once forecasts are evaluated against a strict random-walk benchmark using temporally appropriate data [49].

Several limitations of this study merit acknowledgment. First, the analysis relies exclusively on endogenous predictors (past exchange rate values). The inclusion of macroeconomic fundamentals, interest rate differentials, inflation differentials, commodity prices, and trade balances may improve ML model performance, particularly over longer horizons where fundamental forces exert greater influence [4, 50]. Second, the sample covers a single currency pair. Extending the analysis to other currency pairs, particularly the EUR/USD or GBP/USD, would allow assessment of whether the findings generalize beyond the commodity-currency context. Third, the 18-month test window, while appropriate for the available sample, is shorter than is ideal for robust statistical inference in the DM test framework.

Future research should explore three extensions. First, augmenting the feature set with macroeconomic fundamentals (interest rate differentials, oil price changes, and inflation differentials) may enable ML models to exploit the commodity-currency characteristics of the CAD/USD pair. Second, applying the framework to multiple currency pairs would establish the generalizability of these findings beyond the USD/CAD context. Third, hybrid and stacking approaches that combine ML predictions with economic fundamentals through ensemble averaging or model combination may offer improvements over pure data-driven approaches.

**Declaration of Competing Interest**

The authors declare that they have no known competing financial interests or personal relationships that could have appeared to influence the work reported in this paper.

**Author Contributions**

All authors declare to have contributed equally to this manuscript. All authors read and approved of the final manuscript submitted for publication.

**Data Availability**

The data used to support the findings of this study are publicly available and can be accessed from the Bank of Canada's official website (https://www.bankofcanada.ca).

**Acknowledgment**

Edmund Fosu Agyemang acknowledges the enormous support of the Tulane University Dean Research Council Scholarship and the University of Texas Rio Grande Valley (UTRGV) Presidential Research Fellowship Fund.

**AI Declaration**

The authors declare that AI tools were used solely to improve the language and clarity of the manuscript. All scientific work, tables, and findings were created and validated exclusively by the authors.